\PassOptionsToPackage{table,dvipsnames}{xcolor}
\documentclass{article}

\usepackage{microtype}
\usepackage{graphicx}
\usepackage{subfigure}
\usepackage{booktabs} 

\usepackage{hyperref}

\usepackage[accepted]{icml2025}

\usepackage{amsmath}
\usepackage{amssymb}
\usepackage{mathtools}
\usepackage{amsthm}
\usepackage{xspace}
\usepackage{booktabs}
\usepackage{multirow}
\usepackage{algorithm}
\usepackage{algorithmic}

\usepackage{pifont}
\usepackage{soul}   
\usepackage[normalem]{ulem}   
\usepackage{enumitem}

\usepackage[capitalize,noabbrev]{cleveref}

\theoremstyle{plain}

\theoremstyle{definition}

\theoremstyle{remark}

\newcommand*{\ablate}{\textsc{Ablate}}

\newcommand{\ours}{SelfCite\xspace}
\newcommand{\cmark}{\ding{51}}
\newcommand{\xmark}{\ding{55}}

\usepackage[disable,textsize=tiny]{todonotes}

\icmltitlerunning{SelfCite: Self-Supervised Alignment for Context Attribution in LLMs}

\begin{document}

\twocolumn[
\icmltitle{SelfCite: Self-Supervised Alignment for\\ Context Attribution in Large Language Models}
\begin{icmlauthorlist}
\icmlauthor{Yung-Sung Chuang}{mit}
\icmlauthor{Benjamin Cohen-Wang}{mit}
\icmlauthor{Shannon Zejiang Shen}{mit}
\icmlauthor{Zhaofeng Wu}{mit}
\icmlauthor{Hu Xu}{meta}
\icmlauthor{Xi Victoria Lin}{meta}
\icmlauthor{James Glass}{mit}
\icmlauthor{Shang-Wen Li}{meta}
\icmlauthor{Wen-tau Yih}{meta}
\end{icmlauthorlist}
\icmlaffiliation{mit}{Massachusetts Institute of Technology, Cambridge, MA 02139, USA}
\icmlaffiliation{meta}{Meta FAIR, USA}
\icmlcorrespondingauthor{Yung-Sung Chuang}{yungsung@mit.edu}
\icmlkeywords{Machine Learning, ICML}
\vskip 0.3in
]

\printAffiliationsAndNotice{}

\begin{abstract}
We introduce \ours, a novel self-supervised approach that aligns LLMs to generate high-quality, fine-grained, sentence-level citations for the statements in their generated responses. 
Instead of only relying on costly and labor-intensive annotations, \ours leverages a reward signal provided by the LLM itself through \emph{context ablation}: If a citation is necessary, removing the cited text from the context should prevent the same response; if sufficient, retaining the cited text alone should preserve the same response. 
This reward can guide the inference-time best-of-N sampling strategy to improve citation quality significantly, as well as be used in preference optimization to directly fine-tune the models for generating better citations. 
The effectiveness of \ours is demonstrated by increasing citation F1 up to 5.3 points on the LongBench-Cite benchmark across five long-form question answering tasks. The source code is available at \url{https://github.com/facebookresearch/SelfCite}.
\end{abstract}
\section{Introduction}
\label{introduction}

Assistants built using large language models (LLMs) have become ubiquitous in helping users gather information and acquire knowledge~\citep{chatgpt2023,openai2023gpt4}. For instance, when asked about recent news, an assistant can read through dozens of relevant articles---potentially more than a user could comb through themselves---and use these articles as \emph{context} to provide a clear, specific answer to the user's query. While this ability can greatly accelerate information gathering, LLMs often produce hallucinations—content that sounds plausible but is actually fabricated~\citep{ji2023survey}. Even when provided with accurate context,
models may misinterpret the data or include details that are not supported by the context~\citep{shi2024trusting, chuang2024lookback}.

Although completely eliminating hallucinations remains difficult, existing approaches have sought to enhance the reliability of LLMs by providing 
context attributions--commonly referred to as \emph{citations}--which are fine-grained references to relevant evidences from the context, alongside generated responses for user verification~\citep{menick2022teaching, slobodkin2024attribute, zhang2024longcite}. 
While they have shown promise in generating citations, an outstanding challenge is their reliance on annotated data either from human~\citep{menick2022teaching,slobodkin2024attribute} or costly proprietary APIs~\citep{zhang2024longcite} to train models to generate citations. Collecting annotations can be time-consuming or costly, especially with long-context documents.

To address this challenge, we introduce \ours, a novel alignment approach designed to autonomously enhance the quality of citations generated by LLMs without the need for any annotations in the alignment process. 
Drawing inspiration from model interpretability techniques~\citep{lei2016rationalizing,cohen2024contextcite},
\ours leverages the inherent capabilities of LLMs to provide feedback through \emph{context ablation}—a process to evaluate the necessity and sufficiency of a citation.
If removing the cited text prevents the LLM from assigning high probability to the same response, we can infer that it is \emph{necessary} for the LLM. Conversely, if the response remains highly probable despite removing all context other than the cited text, this indicates that the citation is \emph{sufficient} for the LLM to make the claim. This self-evaluation mechanism enables \ours to calculate a reward signal without relying on the annotation processes.

Building on this intuition, we design a reward that can be cheaply computed by the LLM itself, composed by \emph{probability drop} and \emph{probability hold} in context ablation. 
By integrating this reward function into a best-of-N sampling strategy, \ours achieves substantial improvements in citation quality.
Furthermore, we employ this reward for preference optimization using SimPO~\citep{meng2024simpo}, which not only maintains these improvements but also eliminates the need for the computationally expensive best-of-N sampling. We outperform the previous state of the art on the LongBench-Cite benchmark~\citep{zhang2024longcite} by up to 5.3 points in F1 scores, and showing a promising direction to bootstrap the citation quality from LLMs via self-rewarding.

\begin{figure*}[ht]
\includegraphics[width=\textwidth]{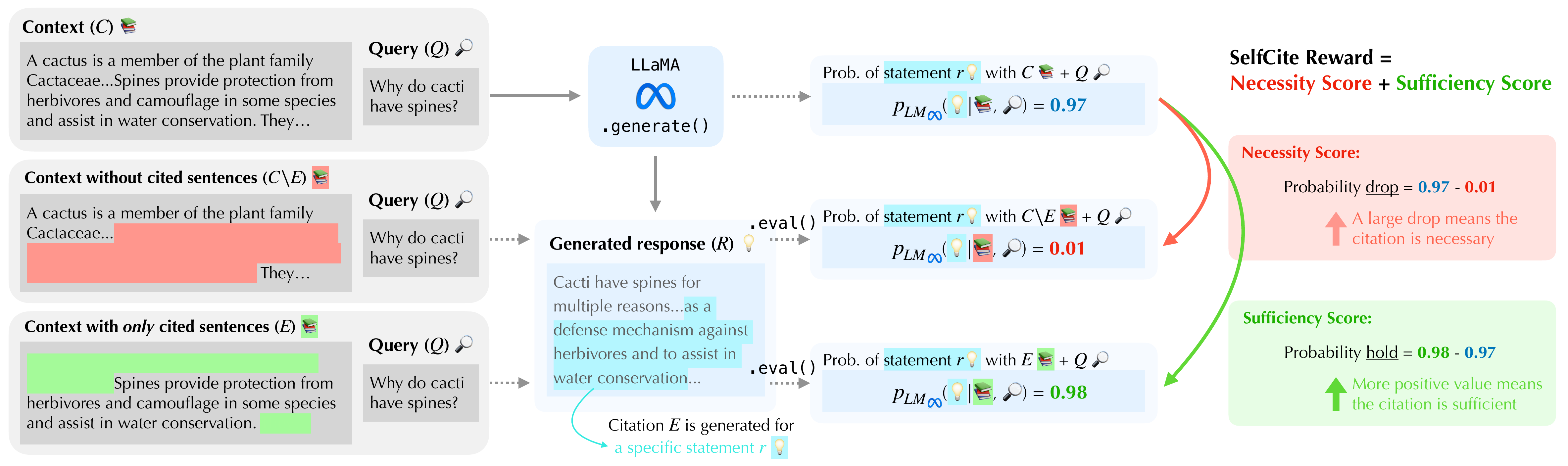}
\vskip -0.15in
\caption{The \ours framework calculates rewards based on two metrics: \emph{necessity score} (probability drop) and \emph{sufficiency score} (probability hold). First, the full context is used to generate a response. Then, the framework evaluates the probability of generating the same response after (1) removing the cited sentences from the context and (2) using only the cited sentences in the context. The probability drop and hold are computed from these probability differences, and their sum is used as the final reward.}
\label{main-figure}
\end{figure*}
\section{Method}
\label{method}

In this section, we describe the \ours framework.
We begin by introducing the task of generating responses with context attributions (\ref{sec:problem_formulation}), referred to as \emph{citations} for brevity.
We then design a reward for providing feedback on citation quality \emph{without} human annotations (\ref{sec:reward}) as illustrated in Fig.~\ref{main-figure}.
Finally, we discuss two approaches for utilizing this reward to improve citation quality: best-of-N sampling (\ref{sec:best_of_n_sampling}) and preference optimization (\ref{sec:preference_optimization}).
\subsection{Problem Formulation}
\label{sec:problem_formulation}

We first formalize the task of generating responses with context attributions and the metrics to self-evaluate context attributions within the \ours framework, inspired by previous papers~\citep{zhang2024longcite, cohen2024contextcite} but adapted to our proposed self-supervised reward.

\paragraph{Setup.} Consider employing an autoregressive language model (LM) to generate a response to a specific query given a context of relevant information. 
Specifically, given an LM $p_{\text{LM}}$, let $p_{\text{LM}}(t_i \mid t_1, \hdots, t_{i-1})$ denote its output distribution over the next token $t_i$ based on a sequence of preceding tokens $t_1,\hdots,t_{i-1}$.
Next, let $C$ represent the context of relevant information.
This context is partitioned into $|C|$ sentences: $c_1, c_2, \dots, c_{|C|}$.
Each sentence $c_j$ is prepended with a unique identifier (e.g., sentence index $j$) as a way for the model to reference the sentence when generating citations. 
The context $C$ is followed by a query $Q$, a question or instruction for the model.
A response $R$ is then sampled from the model $p_\text{LM}$.

\paragraph{Generating Responses with Context Attributions.}

In \ours, following prior work on generating responses with context attributions~\citep{zhang2024longcite}, each statement $r_i$ in the response $R$ is followed by a citation sequence $e_i$ consisting of the identifiers of sentences from the context $C$.
Thus, the entire response sequence $R$ is $\{r_{1}, e_{1}, r_{2}, e_{2}, \dots, r_{S}, e_{S}\}$, where $S$ is the total number of generated statements.
The citation $e_i$ is intended to reference sentences that support the generation of $r_i$. Formally, for each response statement $r_i$, the model outputs a citation sequence $e_i = \{e_{i}^{1}, e_{i}^{2}, \dots, e_{i}^{m}\}$, where each $e_{i}^{j} \in \{1, 2, \dots, |C|\}$ corresponds to a specific sentence number in the context $C$, and $m$ sentences are cited.
Note that this citation sequence may be empty.
The entire response $R$ consisting of statements $r_i$ followed by citations $e_i$ is sampled from the LM $p_\text{LM}$ as follows:
\begin{align*}
r_i &\sim p_{\text{LM}}\left(\cdot \mid c_1, \hdots, c_{|C|}, Q, r_1, e_1, \hdots, r_{i-1}, e_{i-1}\right), \\
e_i &\sim p_{\text{LM}}\left(\cdot \mid c_1, \hdots, c_{|C|}, Q, r_1, e_1, \hdots, r_{i-1}, e_{i-1}, r_{i}\right).
\end{align*}

The objective of optimizing the LM is to ensure that the citation sequence $e_i$ accurately reflects the evidence from the context that supports the generation of $r_i$. 
In the SFT setting~\citep{zhang2024longcite}, the probability of a ``ground truth'' annotated responses and citations $\{\hat{r}_1, \hat{e}_1, ..., \hat{r}_{S}, \hat{e}_{S}\}$ will be maximized, given the input $C$ and $Q$, but it is not trivial to do further alignment with feedback after the SFT data is used up. 
To achieve this, we introduce \ours that can evaluate the quality of these citations based on context ablation as a reward for further preference optimization.

\subsection{Self-Supervised Reward via Context Ablation} 
\label{sec:reward}

We measure the quality of a citation sequence $e_i$ by the \emph{changes} in the LM’s probability of generating $r_i$ when the cited sentences are either removed from or isolated within the context. To simplify the notation, let all the cited context sentences be $E_i = \{c_{e_i^1}, c_{e_i^2}, \dots, c_{e_i^m}\}$. We define two key metrics: \emph{necessity score} and \emph{sufficiency score}, and finally combine them into the final reward, as shown in Fig.~\ref{main-figure}.

\paragraph{Necessity Score: Probability Drop.} This metric quantifies the decrease in the probability of generating $r_i$ when the cited sentences in $E_i$ are all removed from the context (denoted as set minus $\setminus$ operator). Formally, it is defined as:
\[
\text{Prob-Drop}(e_i) = \log p_{\text{LM}}(r_i \mid C) - \log p_{\text{LM}}\left(r_i \mid C \setminus E_i\right).
\]
To keep the equation concise, we ignore $Q$ and $\{r_1, e_1, ..., r_{i-1}, e_{i-1}\}$ in the equation, but they are staying in the context history when computing the probabilities.
A larger probability drop indicates that the removal of $E_i$ significantly diminishes the likelihood of generating $r_i$, thereby validating the necessity of the cited evidence.

\paragraph{Sufficiency Score: Probability Hold.} Conversely, this metric measures if the probability of generating $r_i$ is still kept large when \emph{only} the cited sentences are kept in the context, effectively testing the sufficiency of the citation to support the response statement. Formally:
\[
\text{Prob-Hold}(e_i) = \log p_{\text{LM}}\left(r_i \mid E_i\right) - \log p_{\text{LM}}(r_i \mid C).
\]
A more positive value of probability hold indicates that the cited sentences alone are sufficient to support the generation of $r_i$, while removing all the other irrelevant context. Please note that the values of probability drop or hold can be either positive or negative. For example, if the citation is not relevant to $r_i$ or even distracting, it is possible for $p(r_i \mid E_i)$ to be lower than $p(r_i \mid C)$.

\paragraph{Final Reward.} To comprehensively evaluate the necessity and sufficiency of the generated citations, we add the two metrics together, where the opposing terms cancel out:
\begin{align}
\label{eq:reward}
\text{Reward}(e_i) = \text{Prob-Drop}(e_i) + \text{Prob-Hold}(e_i) \nonumber \\
 = \log p_{\text{LM}}\left(r_i | E_i\right) - \log p_{\text{LM}}\left(r_i | C \setminus E_i\right).
\end{align}
The combined reward measures if the citations are both necessary and sufficient for generating the response $r_i$. 

\subsection{Best-of-N Sampling}
\label{sec:best_of_n_sampling}

To leverage the self-supervised reward computed via context ablation, we employ a \emph{best-of-N} sampling strategy, which is a common way to test the effectiveness of a reward design~\citep{gao2023scaling,lightman2024lets} as a performance oracle without any confounders from training. For convenience, we first generate the full response $R = \{r_1, e_1, \dots, r_{S}, e_{S}\}$ which includes a set of statements ($r_i$) paired with citations ($e_i$), and then locate the position of $e_i$, i.e., where the citation tags \texttt{<cite>...</cite>} are generated. Within the citation tags of $e_i$, we re-sample $N$ candidate citation sequences ($e_i^{(1)}, \dots, e_i^{(N)}$), by making the model to continue the generation from $\{C, Q, r_1, e_1, \dots, r_i\}$, and then select the best citation ($e_i^*$) that maximizes the combined reward metric,~Eq.~\eqref{eq:reward}. The corresponding procedure is shown in Algorithm~\ref{alg:best_of_n}. After obtaining all the selected citations $\{e_1^*, \dots, e_{S}^*\}$, we replace the original citation sequence $e_i$ with the optimal citation $e_i^*$ for each response statement $r_i$, while keeping the response statements $\{r_1, \dots, r_{S}\}$ unchanged. This process is repeated for each statement in the response $R$ to obtain the final, citation-improved output $R^* = \{r_1, e_1^*, \dots, r_{S}, e_{S}^*\}$. To prevent the model from citing too many sentences, we exclude the candidate $e_i$ if the cited text ($E_i$) is longer than $L_\text{max} = 384$ tokens in total, unless $E_i$ are all from a single long sentence.
\vspace{-5pt}
\begin{algorithm}
\caption{\ours Best-of-N Sampling for Citations}
\label{alg:best_of_n}
\begin{algorithmic}
\REQUIRE LM $p_{\text{LM}}$, context $C$, query $Q$, response $R$, \# of candidates $N$, length limit $L_\text{max}$, $T(\cdot)$ counts \# of tokens in a text, $\text{\#}(\cdot)$ counts \# of sentences in a citation.
\FOR{$r_i \in R$}
  \STATE $\mathrm{Reward}(k) = -\infty \text{ for } k = 1, \dots, N$
  \FOR{$k=1,\dots,N$}
    \STATE $e_i^{(k)} \sim p_{\text{LM}}(\cdot \mid r_i, C, Q)$
    \IF{$T(E_i^{(k)})<=L_{\text{max}} \text{ or } \text{\#}(e_i^{(k)})=1$}
        \STATE $\mathrm{Reward}(k)$ \\
        $ = \log p_{\text{LM}}\bigl(r_i | E_i^{(k)}\bigr) 
    - \log p_{\text{LM}}\bigl(r_i | C \setminus E_i^{(k)}\bigr)$
    \ENDIF
  \ENDFOR
  \STATE $k^* = \arg\max_{k}\;\mathrm{Reward}(k)$
  \STATE $e_i^* = e_i^{(k^*)}$
\ENDFOR
\STATE \textbf{return} $R^* = \{r_1, e_1^*, \dots, r_{S}, e_{S}^*\}$
\end{algorithmic}
\end{algorithm}

\subsection{Preference Optimization}
\label{sec:preference_optimization}

Best-of-N sampling is a straightforward way to obtain better citations, but at the additional inference cost of generating candidates and reranking. Thus, we try to internalize the ability of generating better citations back to the LM itself.

Given documents and queries, we can prompt the LM to generate the responses along with the citations $R = \{r_1, e_1, ..., r_{S}, e_{S}\}$. By further applying best-of-N sampling, we can obtain new responses of the same statements but with better citations $R^* = \{r_1, e^*_1, ..., r_{S}, e^*_{S}\}$. Such preference data can be used in direct preference optimization (DPO)~\citep{rafailov2024direct} to align the model based on the preference between the original outputs and improved outputs.
Instead of using DPO, we choose its variant SimPO~\citep{meng2024simpo} here, as SimPO does not require a reference model and allows 2$\times$ memory saving for 25.6K long-context fine-tuning. Through this self-supervised process, which does not require ground-truth answers or human annotations, the model learns to generate more accurate and contextually grounded citations on its own.

\section{Experiments}

We evaluate the effectiveness of \ours by applying the best-of-N sampling and preference optimization methods to existing models that generate responses with citations.

\subsection{Model Details}
\label{subsec:model}

We use LongCite-8B, the Llama-3.1-8B model~\citep{dubey2024llama} fine-tuned on LongCite-45K SFT data~\citep{zhang2024longcite} as the start point 
for both best-of-N sampling and preference optimization. We adopt the same text segmentation strategy from \citet{zhang2024longcite}: each document is split into individual sentences using NLTK~\citep{bird2006nltk} and Chinese punctuations, and each sentence is prepended with a unique identifier in \texttt{\small <C\{$i$\}>} format. These identifiers serve as the \emph{citation indices}, enabling the model to cite relevant context right after the statements with the format of \texttt{\small <statement> \{content ...\} <cite>[}$i_1-i_2$\texttt{\small ][}$i_3-i_4$\texttt{\small ]...</cite></statement>}. This format allows the model to cite a single sentence (e.g. $i_1 = i_2$) or a span (e.g. $i_1 < i_2$) efficiently within several tokens. The responses are generated via top-p sampling~\citep{Holtzman2020The} with p=0.7 and temperature=0.95. We set p=0.9 and temperature=1.2 when doing best-of-N sampling for the citation strings to increase the diversity. We set N=10 in all the experiments considering the limited diversity in citations.\footnote{After deduplicating repeated citation candidates, on average there are only 4.8 candidates left per statement in the BoN experiment on LongBench-Cite, with a standard deviation of 3.2.}

\subsection{Preference Optimization}
\label{sec:po}

\paragraph{LongCite-45K.} 

Best-of-N sampling (Section~\ref{sec:best_of_n_sampling}) requires no training, so no training data is used. For preference optimization with SimPO (Section~\ref{sec:preference_optimization}), we use 2K document–question pairs from LongCite-45K~\citep{zhang2024longcite} as the training set but we do not use its ground-truth responses with high-quality citations for SFT. Instead, we generate model responses from the documents and queries, then apply best-of-N to refine citations. We label the original responses as \emph{rejected} and replace their citations with BoN-refined ones to create the \emph{chosen} responses, forming preference pairs to build the dataset for SimPO.

\paragraph{Data Construction and Length Balancing}

Since best-of-N responses tend to have slightly longer citations, directly fine-tuning on them can lead the model to adopt a shortcut—generating longer citations instead of improving citation quality. To prevent this, we introduce \emph{length balancing}: if an original response has a shorter citation length than the best-of-N response, we insert random citations from nearby sentences. This encourages the model to focus on \emph{where} to cite rather than simply citing \emph{more}. Details are provided in Appendix~\ref{appx:length}, with an ablation study in Section~\ref{sec:balance}.

\begin{table*}[t]
	\vskip -0.2 in
	\caption{Citation recall (R), citation precision (P), citation F1 (F1), and citation length evaluated on LongBench-Cite benchmark. The best of our results are bolded. The best of previous state of the art are underlined. $^\dagger$ indicates the results taken from~\citet{zhang2024longcite}.
	}
	\label{tab:main_cite}
	\centering
	\small
	\resizebox{\linewidth}{!}{
		\setlength{\tabcolsep}{5pt}
		\begin{tabular}{l|ccc|ccc|ccc|ccc|ccc|c|c}
			\toprule
			\multirow{2}{*}{\bf Model}           & \multicolumn{3}{c|}{\bf Longbench-Chat} & \multicolumn{3}{c|}{\bf MultifieldQA} & \multicolumn{3}{c|}{\bf HotpotQA} & \multicolumn{3}{c|}{\bf Dureader} & \multicolumn{3}{c|}{\bf GovReport} & \bf Avg. & \bf Citation                                                                               \\ 
			                                 & R                                       & P                                     & F1                                & R                                 & P                                  & F1       & R            & P    & F1   & R    & P    & F1   & R    & P    & F1   & \bf F1 & \bf Length \\ \midrule
			\multicolumn{18}{l}{\textit{\bf Proprietary models}}                                                                                                                                                                                                                                                                                    \\\midrule
			GPT-4o$^\dagger$                 & 46.7                                    & 53.5                                  & 46.7                              & \underline{79.0}                  & 87.9                               & 80.6     & 55.7         & 62.3 & 53.4 & 65.6 & 74.2 & 67.4 & 73.4 & 90.4 & 79.8 & 65.6   & 220        \\
			Claude-3-sonnet$^\dagger$        & 52.0                                    & 67.8                                  & 55.1                              & 64.7                              & 85.8    & 71.3 & 46.4 & 65.8 & 49.9 & 67.7 &\underline{89.2} & \underline{75.5} & 77.4 & \underline{\bf 93.9} & 84.1 & 67.2   & 132        \\
			GLM-4$^\dagger$                  & 47.6                                    & 53.9                                  & 47.1                              & 72.3           & 80.1                               & 73.6 & 47.0 & 50.1 & 44.4 & 73.4 & 82.3 & 75.0 & \underline{82.8} & 93.4 &        \underline{87.1} & 65.4   & 169        \\ \midrule
			\multicolumn{18}{l}{\textit{\bf Open-source models }}                                                                                                                                                                                                                                                                                   \\\midrule
			GLM-4-9B-chat$^\dagger$          & 25.9                                    & 20.5                                  & 16.7                              & 51.1                              & 60.6                               & 52.0     & 22.9         & 28.8 & 20.1 & 45.4 & 48.3 & 40.9 & 5.7  & 8.2  & 6.3  & 27.2   & 96         \\
			Llama-3.1-8B-Instruct$^\dagger$  & 14.1                                    & 19.5                                  & 12.4                              & 29.8                              & 44.3                               & 31.6     & 20.2         & 30.9 & 20.9 & 22.0 & 25.1 & 17.0 & 16.2 & 25.3 & 16.8 & 19.7   & 100        \\
			Llama-3.1-70B-Instruct$^\dagger$ & 25.8                                    & 32.0                                  & 23.2                              & 53.2                              & 65.2                               & 53.9     & 29.6         & 37.3 & 28.6 & 38.2 & 46.0 & 35.4 & 53.4 & 77.5 & 60.7 & 40.4   & 174        \\
			Mistral-Large-Instruct$^\dagger$ & 19.8                                    & 23.9                                  & 19.0                              & 71.8                              & 80.7                               & 73.8     & 34.5         & 40.9 & 32.1 & 58.3 & 67.0 & 60.1 & 67.9 & 79.6 & 72.5 & 51.5   & 132        \\
			\midrule

			\multicolumn{18}{l}{\textit{\bf Contributive context attribution} (\textit{with Llama-3.1-8B-Instruct})}                                                                                                                                                                                                                                \\\midrule
			ContextCite (32 calls)           & 56.7                                    & 76.8                                  & 58.0                              & 76.1                              & 87.2                               & 78.9     & 40.5         & 54.7 & 43.9 & 58.0 & 82.4 & 65.0 & 67.1 & 88.8 & 75.6 & 64.3   & 92.7       \\
			ContextCite (256 calls)          & \underline{63.5}                        & \underline{\bf 83.1}                  & 64.7                              & 78.8                              & 89.8                       & \underline{81.8} & 46.5         & 60.8 & 49.2 & 61.7 & 89.1 & 70.1 & 69.1 & 93.5 & 78.8 & 68.9   & 100.8      \\
			\midrule

			\multicolumn{18}{l}{\textit{\bf Fine-tuned models}}                                                                                                                                                                                                                                                                                     \\\midrule
			LongCite-9B$^\dagger$            & 57.6                                    & 78.1                                  & 63.6                              & 67.3        & 91.0      & 74.8 & \underline{61.8}  & \underline{78.8} & \underline{64.8} & 67.6           & \underline{89.2} & 74.4 & 63.4 & 76.5 & 68.2 & 69.2   & 91         \\
			LongCite-8B$^\dagger$            & 62.0                                    & 79.7                                  & \underline{67.4}                  & 74.7                              & \underline{93.0}       & 80.8     & 59.2 & 72.1 & 60.3 & \underline{68.3} & 85.6 & 73.1 & 74.0 & 86.6 & 78.5 &\underline{72.0}& 85         \\ [-1em]
   \multicolumn{18}{c}{\tikz[baseline]{\draw[dashed] (0,-0.09cm) -- (19.3cm,-0.09cm);}}\\ [0.1em]
            
                + \textit{SimPO w/ NLI Rewards} & 64.4 & 87.1 & 69.8 & 70.1 & 92.4 & 77.4 & 58.8 & 78.1 & 63.2 & 69.4 & 91.1 & 77.2 & 83.7 & 93 & 87.5 & 75.0 & 105.9 \\ \midrule
			\multicolumn{18}{l}{\textit{\bf Ours: \ours}}                                                                                                                                                                                                                                                                                           \\\midrule

			LongCite-8B (Our repro.)         & 67.0                                    & 78.1                                  & 66.6                              & 74.8                              & 90.7                               & 79.9     & 60.8         & 77.9 & 64.1 & 67.1 & 87.2 & 73.7 & 81.6 & 89.3 & 84.5 & 73.8   & 83.5       \\
			+ BoN                            & 68.4                                    & 81.3                                  & 71.2                              & 76.1                              & 92.8                               & 81.2     & 67.2 & 81.0   & 68.8 & 70.6 & 90.9 & 76.9 & \bf 87.6 & 92.4 & \bf 89.3 & 77.5   & 93.4       \\
			+ SimPO                          & 68.1                                    & 79.5                                  & 69.1                              & 75.5                              & 92.6                               & 81.0     & \bf 69.4 & 82.3 & \bf 71.5 & 72.7 & 91.6 & 78.9 & 86.4 & 92.9 & 89.1 & 77.9   & 105.7      \\
			+ SimPO then BoN                 & \bf 73.3                                & 79.4                                  & \bf 72.8                          & 76.7                              & \bf 93.2    & 82.2 & \bf 69.4 & \bf 83.0 & 71.1 & \bf 74.2 & \bf 92.2 &                \bf 80.3 & 86.7 & 92.7 & 89.2 & \bf 79.1   & 94.7       \\

			\midrule
			\multicolumn{18}{l}{Llama-3.1-8B-Instruct (\textit{fully self-supervised setting})}                                                                                                                                                                                                                                                                                                         \\
			+ SFT on ContextCite             & 52.3                                    & 70.6                                  & 56.5                              & 79.1                              & 90.5                               & 82.0     & 54.5         & 72.3 & 56.3 & 54.9 & 79.0 & 61.6 & 63.7 & 84.9 & 72.3 & 65.7   & 83.0       \\
			\hspace{4mm} + BoN               & 54.8                                    & 67.6                                  & 58.1                              & 80.4                              & 90.5                               & 83.0     & 58.3         & 70.0 & 57.5 & 57.6 & 79.0 & 63.1 & 67.2 & 84.8 & 74.6 & 67.3   & 80.4       \\
			\hspace{4mm} + SimPO             & 63.3                                    & 74.3                                  & 64.6                              & 80.2                              & 88.9                               & 82.4     & 59.7         & 76.9 & 61.0 & 59.0 & 80.9 & 65.4 & 68.5 & 86.6 & 76.1 & 69.9   & 90.2       \\
			\hspace{4mm} + SimPO then BoN    & 66.0                                    & 82.4                                  & 71.1                              & \bf 81.5                          & 90.7                               & \bf 83.2 & 61.3         & 70.0 & 59.9 & 62.1 & 81.4 & 67.4 & 68.8 & 86.2 & 76.1 & 71.5   & 87.4       \\
			\bottomrule
		\end{tabular}
	}
	\vskip -0.1 in
\end{table*}

\subsection{Evaluation}
\label{subsec:datasets}

\paragraph{Benchmark.}
We evaluate our approach on \textbf{LongBench-Cite}~\citep{zhang2024longcite}, a comprehensive benchmark specifically designed for \emph{long-context QA with citations (LQAC)}. Given a long context $C$ and a query $Q$, the model must produce a multi-statement answer with each statement cites relevant supporting sentences in $C$. 
Unlike chunk-level citation schemes~\citep{gao2023enabling} which cites short paragraphs, LongBench-Cite adopts \emph{sentence-level} citations to ensure semantic integrity and finer-grained evidence tracking.
LongBench-Cite assesses two main aspects:
\begin{itemize}[noitemsep,nolistsep]
\item \textbf{Citation Quality:} Whether each statement is fully supported by relevant and \emph{only} relevant sentences. GPT-4o measures \emph{citation recall} (extent to which a statement is fully or partially supported by the cited text) and \emph{citation precision} (whether each cited text truly supports the statement). These are combined into a \emph{citation F1} score. Additionally, we track \emph{average citation length} (tokens per citation) to promote fine-grained citations over unnecessarily long passages.
\item \textbf{Correctness:} How accurately and comprehensively the response answers the query disregarding the citations. This is scored by GPT-4o in a zero-/few-shot fashion based on the query and reference answers.
\end{itemize}

The benchmark contains five datasets, including single-doc QA \textit{MultiFieldQA-en/zh}~\citep{bai2023longbench}, multi-doc QA \textit{HotpotQA}~\citep{yang2018hotpotqa} and \textit{DuReader}~\citep{he2018dureader}, one summarization dataset \textit{GovReport}~\citep{huang2021efficient}, and \textit{LongBench-Chat}~\citep{bai2024longalign} which covers diverse real-world queries with long contexts such as document QA, summarization, and coding.

\paragraph{Baselines.} \ours is compared with these baselines.
\begin{itemize}[noitemsep,nolistsep]
    \item \textbf{Prompting}: \citet{zhang2024longcite} propose the baseline of prompting LLMs with an one-shot example. This can be applied to proprietary models including GPT-4o~\citep{openai2023gpt4}, Claude-3-sonnet~\citep{claude-3}, and GLM-4~\citep{glm2024chatglm}, as well as open-source models including GLM-4-9B-chat~\citep{glm2024chatglm}, Llama-3.1-\{8,70\}B-Instruct~\citep{dubey2024llama}, and Mistral-Large-Instruct~\citep{mistral}.
    \item \textbf{Contributive context attribution}: 
    Contributive context attribution seeks to directly identify the parts of the context that \emph{cause} the model to generate a particular statement. 
    We consider ContextCite \citep{cohen2024contextcite}, a contributive context attribution method that performs several random context ablations to model the effect of ablating different parts of the context on a generated statement. We use NLTK to split Llama-3.1-8B-Instruct's responses into statements,
    and then apply ContextCite with 32 and 256 times of random context ablations to get the citations, with the details described in Appendix~\ref{appx:cc}.
    \item \textbf{Fine-tuned models}: LongCite-8B and 9B released by \citet{zhang2024longcite}, trained on LongCite-45K, fine-tuned from Llama-3.1-8B~\citep{dubey2024llama} and GLM-4-9B~\citep{glm2024chatglm}, respectively.
    Additionally, we consider a baseline of finetuning LongCite-8B using SimPO with the NLI rewards which resembles \citet{huang2024training}, with the details in Appendix~\ref{appx:nli}.
\end{itemize}

\subsection{Main Results}
\paragraph{Citation Quality.}
Table~\ref{tab:main_cite} presents our main results. Our best-of-N sampling (BoN) consistently improves both \textbf{citation recall} and \textbf{citation precision} across tasks, increasing the overall F1 score from 73.8 to 77.5. Using SimPO to internalize BoN’s gains—\textbf{eliminating the need for costly BoN sampling}—achieves a similar improvement, with an F1 of 77.9. Applying BoN again to the SimPO fine-tuned model further boosts \textbf{F1 by 5.3 points to 79.1}, the highest across the datasets, suggesting room for further gains.
Our results surpass LongCite-8B/9B at similar citation lengths and outperform proprietary model prompting while producing shorter citations. 

To better contextualize the gains of our proposed reward, we additionally implement a variant of SimPO using NLI-based citation precision/recall rewards from \citet{huang2024training} by using the same training pipeline and initialization as our SimPO, modifying only the reward function (see details in Appendix~\ref{appx:nli}). As shown in row of \textit{SimPO w/ NLI Rewards}, this baseline improves LongCite-8B on 3 out of 5 datasets, but is still consistently outperformed by SelfCite. This result highlights that while NLI-based rewards are helpful, our SelfCite reward provides a more accurate signal for optimizing citation quality.

Besides the fine-tuned baselines, we additionally compare our method to ContextCite for reference, a method very different from SelfCite--it does not directly generate citations, it estimates the importance scores of the context sentences after the response is generated (in Appendix~\ref{appx:cc} we show how to convert continuous importance scores into citations). Both SelfCite and ContextCite rely on the idea of context ablation, but our approach is significantly better. A key reason is that ContextCite estimates sentence importance from scratch using linear regression, while we rerank existing LLM-generated citation candidates, leading to more efficient and accurate citation quality estimation.

Finally, we evaluate the latest released \emph{Claude Citations} API, as shown in Appendix~\ref{appx:claude} that \ours achieves strong results very close to this commercial-level API, validating the effectiveness of SelfCite.

\paragraph{Fully Self-Supervised Setting.}

In our main experiment, we start from the Llama-3.1-8B model fine-tuned on the LongCite-45K SFT data, which effectively kick-starts its ability to generate structured citations for best-of-N sampling. The subsequent SimPO alignment stage is entirely self-supervised. We are also curious if it is possible to start from a fully self-supervised SFT model and then apply our self-supervised alignment after that. To begin with, we automatically generate 11K citation SFT data using ContextCite (see Appendix~\ref{appx:cc} for details) to replace the LongCite-45K annotations in the training data, as shown in the results at the bottom of Table~\ref{tab:main_cite}. We can see that SFT on ContextCite can achieve decent initial results (65.7 F1) but still far from LongCite-8B (73.8 F1). BoN helps improving F1 to 67.3. After SimPO training, it achieves 69.9 F1, and additionally applying BoN can boost its F1 by 5.8 to 71.5, significantly closing the gap to LongCite-8B, showing our alignment method not only improve the supervised models, but also enhance the models purely trained from self-supervision.

\begin{table}[t]
\vskip -0.1 in
\caption{Answer correctness when responding with or without citations. $^\dagger$ indicates results taken from \citet{zhang2024longcite}. The header contains abbreviations for the same five datasets in Table~\ref{tab:main_cite}.}
\vskip 0.05 in
\label{tab:correctness}
\centering
\small
\resizebox{\linewidth}{!}{
\begin{tabular}{l|c|c|c|c|c|c}
\toprule
 \bf Model 
 & \bf Long.
 & \bf Multi.
 & \bf Hot.
 & \bf Dur.
 & \bf Gov.
 & \bf Avg \\ 
\midrule
\multicolumn{7}{l}{\textit{Answering without citations}} \\
\midrule
LongSFT-8B$^\dagger$              & 68.6 & 83.6 & 69.0 & 62.3 & 54.4 & 67.6 \\
LongSFT-9B$^\dagger$              & 64.6 & 83.3 & 67.5 & 66.3 & 46.4 & 65.6 \\ [0.1em]
Llama-3.1-8B-Instruct   & 66.0 & 83.7 & 65.8 & 62.8 & 66.1 & 68.9 \\
\midrule
\multicolumn{7}{l}{\textit{Answering with citations}} \\
\midrule
LongCite-8B (Our repro.)      & 67.6 & 86.7 & 69.3 & 64.0 & 60.4 & 69.6 \\
+ SimPO                       & 67.4 & 86.7 & 67.5 & 66.0 & 61.3 & 69.8 \\
\midrule
Llama-3.1-8B-Instruct         & 58.4 & 75.3 & 67.3 & 59.3 & 56.4 & 63.3 \\
+ SFT on ContextCite          & 58.8 & 83.4 & 65.8 & 57.8 & 57.5 & 64.6 \\
\hspace{4mm} + SimPO          & 56.8 & 80.9 & 65.3 & 59.5 & 60.9 & 64.7 \\
\bottomrule
\end{tabular}
}
\vskip -0.2in
\end{table}

\paragraph{Answer Correctness.}

For best-of-N sampling, only the citation parts are modified, so the responses it generates to answer the questions are the same as those of the original LongCite-8B model, maintaining the same correctness. For the SimPO fine-tuned models, we test their answer correctness by the evaluation in \citet{zhang2024longcite}, which contains two settings: answering with/without citations. If answering with citations, the model will be prompted to generate answers with structured citations, making the task more complex, and the citation parts will be removed when evaluating the answer correctness. 
The results in Table~\ref{tab:correctness} show that the SimPO fine-tuning \textbf{does not change the correctness} of the LongCite-8B model much. The correctness is similar to LongSFT-8B/9B~\citep{zhang2024longcite}, which are ablation baselines fine-tuned on LongCite-45k QA pairs but without the citation parts. 
The same observation still holds when starting from Llama-3.1-8B-Instruct, either SFT with ContextCite data or the further SimPO step do not change the answer correctness significantly. Under the same answer correctness, the additional “citations” can benefit the verifiability of the answers, enabling a user to easily double-check the answer, even in cases where the answers are wrong.

\paragraph{Chunk-level Citation Evaluation.} Additionally, we evaluate our methods on the traditional chunk-level citation benchmark \textbf{ALCE}~\citep{gao2023enabling}. However, due to the mismatch of data distributions and different task settings during training (sentence-level) and evaluation (chunk-level), we consider this as a zero-shot evaluation, and the results are shown in Appendix~\ref{appx:alce}, due to the limited space.
\section{Analysis}
\subsection{Ablation Study on Rewards}
\label{sec:ablation}

To better understand our final reward design, we explore various reward strategies in the BoN sampling process. Here, all BoN candidates are pre-generated and fixed, the reward is the only factor affecting results. Table~\ref{tab:ablation_decoding} presents our ablation results on HotpotQA, while citation lengths are computed across all LongBench-Cite datasets for direct comparison with Table~\ref{tab:main_cite}.
We evaluate four alternative reward designs. \textit{BoN by LM log prob} re-ranks candidates simply by the probability of the citation string, \texttt{\small <cite>[}$i_1-i_2$\texttt{\small ][}$i_3-i_4$\texttt{\small ]...</cite>}, which is similar to beam search but less costly. We observe that this strategy slightly boosts recall while reducing precision, resulting in a minor reduction in F1. 
\textit{BoN by max citation length} always selects the candidates with the longest citations, i.e. citing the greatest number of sentences. Although it improves recall, it significantly reduces precision from 77.9 to 73.6 and inflates the citation length from 83.5 to 139.8. 
By contrast, both \textit{BoN by Prob-Drop} and \textit{BoN by Prob-Hold} improve recall without sacrificing precision. 
Finally, by combining both Prob-Drop and Prob-Hold into our final \ours reward, we achieve the best outcome, increasing \textbf{both recall and precision and a 4-point improvement in F1}.

We also explored different token-length limits for citations in the bottom of Table~\ref{tab:ablation_decoding}, as discussed in \Cref{sec:best_of_n_sampling}. By default, we exclude candidates citing more than 384 tokens, unless the citation contains only a single sentence. Lowering the cap to 256 tokens slightly hurts F1, while raising it to 512 tokens has negligible impact. Completely removing length limits inflates citation length to 121.9 tokens and yields worse precision (79.3) but slightly improved recall (67.9). We also notice that the 256 length limit still outperforms the LongCite-8B baseline (66.4 vs 64.1) while having almost equally long citation length (84.5 vs 83.5), showing that \textbf{the improvement of \ours correlates less with the citation length}.
Overall, using a 384-token limit achieves a good balance for short citation lengths and strong performance.
    \begin{table}[t]
    \vskip -0.1 in
    \caption{Ablation study on HotpotQA citation recall, precision, and F1 (R, P, F1) and citation length for BoN decoding methods.}
    \label{tab:ablation_decoding}
    \vskip 0.05 in
    \centering
    \small
    \resizebox{0.475\textwidth}{!}{
    \begin{tabular}{l|ccc|c}
    \toprule
    \multirow{2}{*}{\bf Decoding Methods} & \multicolumn{3}{c|}{\bf HotpotQA} & \bf Citation \\
     & \bf R & \bf P & \bf F1 & \bf Length \\
    \midrule
    LongCite-8B (Our repro.) & 60.8 & 77.9 & 64.1 & 83.5 \\
    \midrule
    + BoN by LM log prob     & 62.7	& 75.5 & 63.4 & 74.6  \\
    + BoN by \textit{max citation length}   & 66.5 & 73.6 & 65.1 & 139.8 \\
    + BoN by Prob-Drop   & 65.6 & 78.1 & 66.6 & 92.9 \\
    + BoN by Prob-Hold   & 66.2 & 78.1 & 67.0 & 93.4 \\
    \midrule
    + BoN by \ours   & 67.2 & 81.0 & 68.8 & 93.4 \\[-0.9em]
    \multicolumn{5}{c}{\tikz[baseline]{\draw[dashed] (0,-0.06cm) -- (8.5cm,-0.06cm);}} \\[0.1em]
    {   w/} lower length limit (256) & 65.8 & 78.8 & 66.4 & 84.5 \\
    {   w/} higher length limit (512) & 67.0 & 82.2 & 68.5 & 99.2 \\
    {   w/o} length limit ($\infty$) & 67.9 & 79.3 & 68.1 & 121.9 \\
    \bottomrule
    \end{tabular}
    }
    \vskip -0.1 in
    \end{table}

    \begin{table}[t]
    \vskip -0.1 in
    \caption{Ablation study on HotpotQA citation recall, precision, and F1 (R, P, F1) and citation length for finetuned models.}
    \label{tab:ablation_finetuning}
    \vskip 0.05 in
    \centering
    \small
    \resizebox{0.45\textwidth}{!}{
    \begin{tabular}{l|ccc|c}
    \toprule
    \multirow{2}{*}{\bf Fine-tuning Methods} & \multicolumn{3}{c|}{\bf HotpotQA} & \multicolumn{1}{c}{\bf Citation} \\
    & \bf R & \bf P & \bf F1 & \bf Length \\
    \midrule
    LongCite-8B (Our repro.) & 60.8 & 77.9 & 64.1 & 83.5 \\
    \midrule
    + SimPO                  & 69.4 & 82.3 & 71.5 & 105.7 \\
    + SimPO + BoN            & 72.0 & 82.7 & 72.9 & 126.9 \\
    \midrule
    \multicolumn{5}{l}{\it + SimPO w/ or w/o length balancing} \\
    \midrule
     w/ length balancing      & 69.4 & 82.3 & 71.5 & 105.7 \\
     w/o length balancing     & 64.4 & 62.9 & 60.5 & 152.9 \\
    \midrule
    \multicolumn{5}{l}{\it + SimPO w/ varying data sizes} \\
    \midrule
     1K examples              & 62.5 & 78.9 & 65.7 & 90.1 \\
     2K examples              & 69.4 & 82.3 & 71.5 & 105.7 \\
     4K examples              & 68.5 & 80.4 & 70.3 & 134.1 \\
     8K examples              & 64.6 & 79.5 & 65.9 & 158.1 \\
    \midrule
    + {\it SFT on BoN responses}      & 68.8 & 77.3 & 68.4 & 98.7 \\
    \midrule
    \multicolumn{5}{l}{\it + SimPO by denoising perturbed citations} \\
    \midrule
    On original responses & 40.5 & 50.5 & 41.6 & 88.8 \\
    On BoN responses & 42.6 & 50.7 & 42.3 & 79.7 \\
    \bottomrule
    \end{tabular}
    }
    \vskip -0.23 in
    \end{table}

\vspace{-5pt}
\subsection{Citation Length Balance}
\label{sec:balance}

As noted in Section~\ref{sec:po}, BoN selects slightly longer citations, making it easy for a model trained directly on BoN-preferred data to adopt the shortcut of generating longer citations without improving quality. To counter this, we apply \emph{length balancing}, injecting random citations into examples where length bias exists to equalize the number of cited sentences. Table~\ref{tab:ablation_finetuning} (see w/ vs. w/o length balancing) highlights its critical role in length balancing. Without length balancing, the model overextends citations (average length 152.9), leading to lower precision (62.9) and F1 (60.5). In contrast, enabling length balancing maintains high precision (82.3) and recall (69.4), achieving a better F1 of 71.5 while keeping citation length reasonable (105.7). These results confirm that \textbf{length balancing prevents shortcut learning, ensuring the model truly learns to cite accurately}.

\subsection{Training Size of SimPO}

In prior study~\citep{zhou2023lima}, 1K examples are sufficient to align user preferences effectively. Table~\ref{tab:ablation_finetuning} presents SimPO results with 1K to 8K examples. 1K examples already bring a moderate improvement, raising F1 from 64.1 to 65.7, with gains in precision and recall. Using 2K examples further boosts F1 to 71.5, while 4K leads to saturated improvement. However, at 8K examples, performance declines, and citation length rises to 158.1. We attribute this to SimPO’s off-policy nature, especially because it lacks a reference model to constrain the output distributions to be similar to the collected data. As training steps grow, the model may drift from the collected data, potential overfitting to the biases in preference data. Thus, further fine-tuning may degrade citation quality. To address this, we show initial results from iterative SimPO in \Cref{sec:iter}.

\subsection{SimPO vs.\ SFT on Best-of-N responses}

We also show the effect of applying standard supervised fine-tuning (SFT) on the responses selected by best-of-N sampling, which is a simplified alternative of preference optimization. As the result shown in the last row in Table~\ref{tab:ablation_finetuning}, SFT also improves the F1 score from 64.1 to 68.4, but it still falls behind 71.5 of SimPO. This result confirms that it is necessary to train the model via SimPO with preference data, which enables the model to distinguish between bad and good citations, and thus improve the citation quality.

\subsection{Off-policy Denoising Perturbed Citations}
\label{sec:offpolicy}

We explored a purely \emph{off-policy} alternative approach. Specifically, given a model-generated response, we randomly shift its citation spans to create perturbed variants. SimPO training pairs were then constructed by preferring the \emph{original} citation over the \emph{perturbed} one, encouraging the model to ``denoise'' citations by restoring their original spans.
However, as shown at the bottom of Table~\ref{tab:ablation_finetuning}, this approach \emph{degrades} performance, both when applied to original and best-of-N responses. We attribute this to a mismatch between the training data and the model’s natural error distribution—since random shifts do not reflect typical citation errors, they fail to provide useful guidance for improvement.

\subsection{Iterative Preference Optimization}
\label{sec:iter}

\begin{figure}[t!]
\centering
\includegraphics[width=0.9\linewidth]{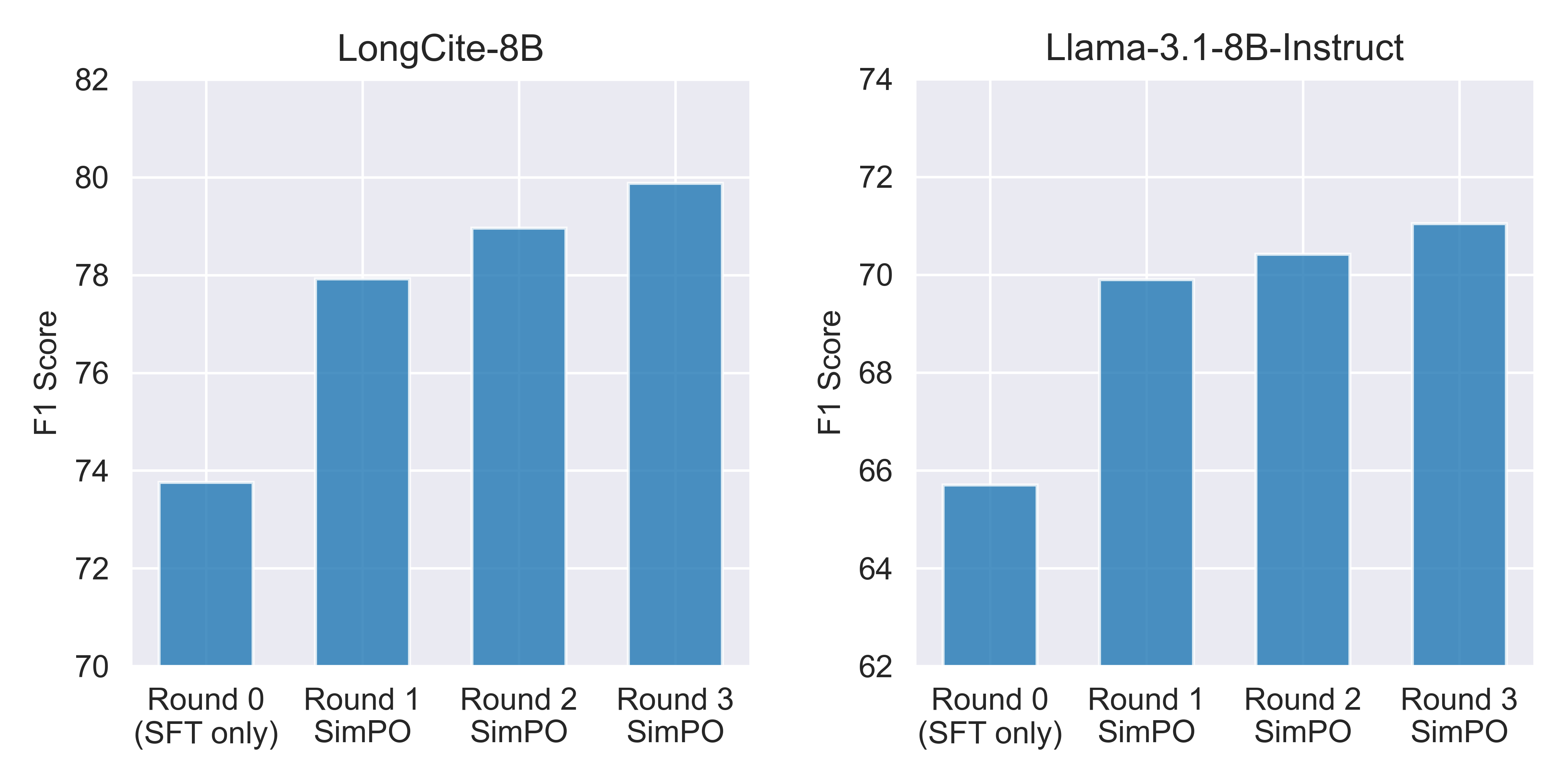}
\vskip -0.15 in
\caption{Iteratively applying SimPO for three iterations.}
\label{fig:iter}
\vskip -0.15 in
\end{figure}

It has been discussed that an \emph{on-policy} alignment process can be beneficial to avoid reward exploitation~\citep{bai2022training} and maintains consistency between the generated data and the model’s evolving output distribution.
We thus experiment with iteratively performing SimPO, similar to the concepts of recent studies~\citep{pang2024iterative, yasunaga2024alma}, to maintain the consistency between the generated data and the model’s evolving output distribution. 
Specifically, after fine-tuning with SimPO, we generate a new dataset via BoN, which is also 2K in size but not overlapped with previous iterations. We continue training the model and repeat the process for three rounds.
As shown in Figure~\ref{fig:iter}, while the largest improvement occurs in the first round, improvements continue over three iterations, which further validates the reliability of our reward signal.
Iterative SimPO is still not perfect since it remains an off-policy method. Given that our reward can be cheaply computed, we believe that on-policy methods like PPO~\citep{schulman2017proximal} could further enhance performance. We leave the exploration of such approaches for future work.

\subsection{Latency of Best-of-N}

Table~\ref{tab:latency} reports the average per-example latency on LongBench-Cite. As expected, Best-of-N (BoN) introduces additional latency due to the need to generate and rerank multiple citation candidates. In our setup, we use N = 10 candidates, but the sampling time is not 10× longer than direct decoding. This is because we only re-sample short citation spans (typically 5–10 tokens), not the full responses, resulting in relatively lightweight sampling overhead.

However, the increased latency from BoN is not a major concern, because our SelfCite SimPO model also achieves the same performance as BoN in a single pass, without additional latency. For scenarios requiring maximum efficiency, we recommend using the SimPO model directly.

\subsection{Qualitative Study}

Finally, we examine an example that requires citing multiple context sentences to support a complex response. As shown in Table~\ref{tab:qual}, the response integrates information from sentences 302, 303, and 306.
Direct sampling (2) omits sentence 302 while incorrectly including 305. In contrast, the best-of-N candidate (1) correctly includes 302 and excludes 305, achieving a slightly higher reward (0.578 vs.\ 0.547), demonstrating the effectiveness of our reward design. We also present candidates (3) and (4), which cite more irrelevant sentences and miss key citations, leading to even lower rewards. Additional qualitative examples are provided in Appendix~\ref{appx:qual}.

\begin{table}[t!]
\centering
\caption{Average latency per example on LongBench-Cite (8 × A100 GPUs, batch size 1, model parallel).}
\vskip 1mm
\begin{tabular}{l c}
\toprule
\textbf{Method} & \textbf{Avg Latency (s)} \\
\midrule
LongCite-8B & 24.3 \\
SelfCite BoN Sampling & 149.0 \\
SelfCite BoN Reranking & 34.0 \\
SelfCite SimPO model & 26.2 \\
\bottomrule
\end{tabular}
\label{tab:latency}
\end{table}

\begin{table*}[t!]
    \vskip -0.1 in
\caption{An example of differences in the citation from baseline vs BoN. Related information are highlighted in the context/response.}
\label{tab:qual}
    \vskip 0.05 in
\centering
\small
\begin{tabular}{p{0.1\textwidth} p{0.85\textwidth}}
\toprule
\bf Sent. ID & \textbf{Context Sentences} (only showing a paragraph due to limited space) \\
\midrule
\textbf{302 (\ding{51})} 
  & In general, consumer advocates believe that any 
    \hl{comprehensive federal privacy policy} 
    should complement, and not supplant, sector-specific \hl{privacy legislation} or state-level legislation.\\
\midrule
\textbf{303 (\ding{51})} 
  & \hl{Finding a global consensus on how to balance open data flows and privacy protection 
    may be key to maintaining trust} 
    \hl{in the digital environment and advancing international trade.}\\
\midrule
\textbf{304 (\ding{55})} 
  & One study found that over 120 countries have laws related to personal data protection.\\
\midrule
\textbf{305 (\ding{55})} 
  & Divergent national privacy approaches raise the costs of doing business and make it harder for governments 
    to collaborate and share data, whether for scientific research, defense, or law enforcement.\\
\midrule
\textbf{306 (\ding{51})} 
  & \hl{A system for global interoperability} in a least trade-restrictive and nondiscriminatory way 
    between \hl{different national} \hl{systems} could help minimize costs and allow entities in different jurisdictions 
    with varying online 
    \hl{privacy regimes} to share data via cross-border data flows.\\
\midrule
\midrule
\bf Query & Please write a one-page summary of the above government report. \\
\midrule
\textbf{Response} (only single statement due to space)
 & [...] The report concludes by noting that 
   \hl{finding a global consensus on how to balance open data flows and privacy protection 
   may be key to maintaining trust in the digital environment and advancing international trade.} 
   The report suggests that 
   Congress may consider \hl{comprehensive privacy legislation} and examine the potential
   challenges and implications of building \hl{a system of interoperability between different national 
   privacy regimes.} [...]
   \\

\multicolumn{2}{l}{
\begin{tabular*}{0.95\textwidth}{@{\extracolsep{\fill}}llll}
\midrule
\textbf{BoN Candidates} & \textbf{Citation Strings (\textcolor{ForestGreen}{\it green: correct}; \textcolor{red}{\it red: wrong})} & \textbf{Missing Citations} & \textbf{\ours Reward} \\
\midrule
(1) Best candidate & 
\texttt{\textcolor{ForestGreen}{[302-303][306-306]}} & 
-- & 
0.578 \\

(2) Direct sampling & 
\texttt{\textcolor{ForestGreen}{[303}\textcolor{ForestGreen}{-303]}\textcolor{ForestGreen}{[}\textcolor{red}{305}\textcolor{ForestGreen}{-306]}} & 
\textcolor{purple}{(302)} & 
0.547 \\

(3) Other candidate & 
\texttt{\textcolor{ForestGreen}{[303-}\textcolor{red}{304}\textcolor{ForestGreen}{]}\textcolor{red}{[308-308][310-311]}} & 
\textcolor{purple}{(302, 306)} & 
0.461 \\

(4) Other candidate & 
\texttt{\textcolor{ForestGreen}{[303-303]}\textcolor{red}{[309-309][311-311]}} & 
\textcolor{purple}{(302, 306)} & 
0.375 \\
\end{tabular*}
}
\\

\bottomrule
\end{tabular}
\end{table*}

\section{Related Work}
\label{sec:related_work}

\paragraph{Citations for Language Models.}

Recent work has explored various approaches to teaching language models to generate citations, including fine-tuning with direct human feedback or annotations~\citep{nakano2021webgpt,menick2022teaching,slobodkin2024attribute}, rewards from external NLI models~\citep{huang2024training,huang2024advancing}, and prompting-based methods~\citep{gao2022rarr, gao2023enabling} to explicitly incorporate relevant retrieved documents. Given the high cost of human annotation, \citet{zhang2024longcite} introduced \textbf{CoF} (“\textbf{Co}arse to \textbf{F}ine”), an automated multi-stage pipeline that simulates human annotation. This approach leverages proprietary LLMs for query generation, chunk-level retrieval, and sentence-level citation extraction, achieving high citation quality through supervised fine-tuning. However, it depends on 
larger proprietary models two proprietary APIs—GLM-4 for the LLM and Zhipu Embedding-v2 for retrieval\footnote{\url{https://open.bigmodel.cn/pricing}}—
with carefully designed prompting, effectively distilling the capabilities of these proprietary models into much smaller models in 8B/9B.
In contrast, our \ours aims at completely eliminating the reliance on annotations for citation, either from human or proprietary APIs. Instead, our method enables a small 8B model to assess citation quality itself using self-supervised reward signal from context ablation, effectively self-improving without external supervision. We additionally provide Table~\ref{tab:method-comparison} to contrast the key differences between SelfCite and prior papers in Appendix~\ref{appx:comparison}.

\paragraph{Contributive Context Attribution.}

Besides being self-supervised, \ours also adopts the view that citations should reference the sources from the context that a model actually \emph{uses} when generating a statement--known as \emph{contributive} attribution~\citep{worledge2023unifying}--rather than any sources that merely \emph{support} the claim.
Our reward signal naturally aligns with this attribution framework, as context ablation identifies the sources that \emph{cause} the model to produce a statement. 
Existing contributive attribution methods for LLMs typically require extensive context ablations or other computationally expensive techniques, such as gradient-based analysis during inference~\citep{cohen2024contextcite,Qi2024ModelIA,phukan2024peering}.
In contrast, \ours simply generate the citation tags, and refine citation candidates by preference optimization with reward signals from context ablations, effectively teaching the model to perform contributive context attribution itself.

We also note that there is a distinction between \emph{corroborative} citation—highlighting sources that \emph{support} a claim, as used in benchmarks like LongBench-Cite—and \emph{contributive} attribution, as emphasized in ContextCite. While SelfCite applies a contributive alignment method (via ablations) in the context of a corroborative evaluation framework, we find the two objectives to be at least partially aligned: citations that genuinely influence the generation are often also semantically supportive. Although this alignment is not guaranteed, our empirical results show that enforcing contributive attribution leads to clear improvements on corroborative benchmarks, suggesting that current corroborative methods (e.g., LongCite) still have significant headroom for improvement—even under a slightly mismatched objective.

\paragraph{Self-Supervised Alignment and Reward Modeling.}
Another relevant area is self- or weakly-supervised approaches for aligning LLMs without human supervision~\citep{kim2023aligning,yuan2024selfrewarding}, reducing the need for explicit human feedback~\citep{ouyang2022training}, or curating high-quality data for supervised fine-tuning~\citep{zhou2023lima}. \ours  shares the same spirit by computing simple probability \emph{differences} under context ablation as rewards, eliminating the need for additional annotation process. 
\section{Conclusion and Limitations}

We present \ours, a self-supervised framework for aligning large language models (LLMs) to generate more accurate and fine-grained citations. By leveraging LLMs' own output probabilities, \ours computes necessity and sufficiency rewards through context ablation, enabling preference optimization without relying on external annotations from human or proprietary APIs. Applying such rewards in best-of-N (BoN) sampling and SimPO fine-tuning can significantly improve the citation correctness on the LongBench-Cite benchmark, offering a promising self-improving direction towards verifiable and trustworthy LLMs.

\ours also has limitations: 1) While achieving strong results with SimPO, integrating other preference optimization or reinforcement learning (RL) algorithms, e.g., PPO~\citep{schulman2017proximal}, remains under explored. However, prior work~\citep{mudgal2024controlled} shows that BoN closely approximates the performance upper bound of RL, and we follow established practice~\citep{gao2023scaling,lightman2024lets} to mainly validate our rewards through BoN, and further verify it with SimPO fine-tuning.
2) \ours assumes access to model output probabilities, which may not be feasible for closed-source models. 3) While our framework improves the quality of citations already generated by LLMs, discovering unsupervised methods to kick-start LLMs' ability in generating structured citations from scratch remains an important direction for future research.

\section*{Impact Statement}

This paper introduces \ours, a self-supervised framework for improving citation accuracy in large language models (LLMs). Our method enhances the verifiability and trustworthiness of LLM-generated content by aligning citations with relevant supporting evidence in a scalable manner, without relying on costly human annotations.
By improving citation quality, \ours contributes to the broader goal of reducing misinformation and hallucinations in AI-generated responses. Ensuring that LLMs provide accurate and properly attributed information is particularly crucial in high-stakes domains such as healthcare, law, and journalism, where incorrect or unverified information can have significant real-world consequences.
Overall, \ours aligns with the broader ethical goal of making machine learning systems more transparent and accountable, reducing the risk of unchecked misinformation while maintaining the efficiency and scalability required for real-world applications.

\section*{Acknowledgements}

We thank Jiajie Zhang and Yushi Bai for their assistance in providing implementation details of LongCite. Special thanks to Pin-Lun (Byron) Hsu for his invaluable support and guidance with Liger-Kernel. We are also grateful to Tianyu Gao for his timely help in setting up the ALCE benchmark during the rebuttal period. We also appreciate Andrei Barbu, Linlu Qiu, Weijia Shi for their valuable discussions. Yung-Sung was sponsored by the Department of the Air Force Artificial Intelligence Accelerator and was accomplished under Cooperative Agreement Number FA8750-19-2-1000. The views and conclusions contained in this document are those of the authors and should not be interpreted as representing the official policies, either expressed or implied, of the Department of the Air Force or the U.S. Government. The U.S. Government is authorized to reproduce and distribute reprints for Government purposes notwithstanding any copyright notation herein.

\bibliography{main,contextcite}
\bibliographystyle{icml2025}

\newpage
\appendix
\onecolumn

\section{Implementation Details}
\label{appx:details}

For SimPO fine-tuning, we randomly sample 2K document and question pairs from the LongCite-45k data, generate the best-of-N responses with our Algorithm~\ref{alg:best_of_n} to obtain the preference data, and train for one epoch. We sample another 100 examples as development set to pick the best learning rate from \{1e-7, 3e-7, 5e-7, 7e-7\}. We keep other hyperparameters the same as the original SimPO~\citep{meng2024simpo}. We follow the same prompt format used in \citet{zhang2024longcite}\footnote{\url{https://github.com/THUDM/LongCite}} to keep the comparison fair.
For the iterative SimPO experiment, in each iteration, we sampled a new, non-overlapping subset of 2K examples to ensure no data repetition across iterations. For self-supervised SFT, we generate 11K citation data unsupervisedly from ContextCite outputs as described in Appendix~\ref{appx:cc}, trained with a larger learning rate 7e-6. 

We use the SimPO source code~\footnote{\url{https://github.com/princeton-nlp/SimPO}} built from Huggingface Transformers~\citep{wolf-etal-2020-transformers} for the finetuning experiments, as well as Liger-Kernel~\citep{hsu2024ligerkernelefficienttriton}\footnote{\url{https://github.com/linkedin/Liger-Kernel}} to enable memory efficient training for long-context examples in LongCite-45K without tensor parallelization. 
We run all the finetuning experiments on with 8$\times$A100 GPUs of 80 GB memory on a single node. The batch size is set to 1 per GPU due to the long context examples.
We set our max context length to 25600 to prevent OOM. For the data examples longer than 25600, we perform truncation, start from truncating the sentences that are the most far away from the sentences cited by the ground truth annotation, so as to keep the impact of truncation to be minimum.

When evaluating the citation length, as well as calculating the token length limit of 384 for excluding long BoN candidates, we follow \citet{zhang2024longcite} to use GLM4-9B’s tokenizer to count tokens.

In the ablation study of off-policy denoising in \Cref{sec:offpolicy}, the citation examples for denoising are collected by randomly shifting existing citation spans by 3-10 positions in sentence indices.

\section{Obtaining Citations from ContextCite}
\label{appx:cc}

In this section, we first describe how the ContextCite method~\citep{cohen2024contextcite} estimates continuous attribution scores for each sentence in the context. We then explain a simple heuristic for extracting citations (i.e., selecting a subset of context sources) from these scores.

\subsection{ContextCite}

Given a language model $p_\text{LM}$, a context $C$, a query $Q$ and a generated response $R$, ContextCite aims to quantify how each \emph{source} in the context $C = \{c_1, c_2, \dots, c_{|C|}\}$ contributes to the generated response $R$ (in our case, the sources are sentences). 
To do so, ContextCite performs several random context ablations.
We begin by introducing some notation to describe these ablations.
Let $v \in \{0,1\}^{|C|}$ be an ablation vector whose $i$-th entry toggles whether source $c_i$ is included ($v_i=1$) or excluded ($v_i=0$).
We write $\ablate(C,v)$ to denote a modified version of the original context $C$ in which sources for which $v_i=0$ are omitted.
ContextCite seeks to understand how the probability of generating the original generated response,
\[
f(v) := p_\text{LM}(R \,\mid\, \ablate(C, v), Q),
\]
changes as a function of the ablation vector $v$.

\paragraph{Attribution via Surrogate Modeling.}
Directly measuring $f(v)$ for all $2^{|C|}$ ablation vectors is infeasible for large $|C|$.
Hence, ContextCite seeks to identify a \textit{surrogate model} $\hat{f}(v)$ that is easy to understand and approximates $f(v)$ well.
To simplify this surrogate modeling task, ContextCite applies a logit transform to $f$, which maps values in $(0,1)$ to $(-\infty,\infty)$):
\[
g(v) := \sigma^{-1}(f(v)) \;=\; \log\!\Bigl(\frac{f(v)}{1 - f(v)}\Bigr).
\]
ContextCite then approximates $g(v)$ using a sparse linear function,
\[
\hat{g}(v) \;=\; \hat{w}^\top v + \hat{b}.
\]
Notice that resulting weights $\hat{w} \in \mathbb{R}^{|C|}$ encode the importance of each source $c_i$ to the probability of generating the original response;
they can be interpreted directly as attribution scores (higher scores suggest greater importance).

\paragraph{Finding a Surrogate Model via \textsc{Lasso}.}

To learn the parameters $\hat{w}$ and $\hat{b}$ of the surrogate model, ContextCite randomly samples a small number of ablation vectors and measures the corresponding probabilities of generating the original response. 
It then uses this ``training dataset'' to fit a sparse linear model with $\textsc{Lasso}$.
Concretely, it learns a surrogate model with the following three steps:
\begin{enumerate}
    \item Sample $n$ ablation vectors $\{v_i\}_{i=1}^n$ uniformly at random from $\{0,1\}^{|C|}$.
    \item For each sample $v_i$, compute $g(v_i) = \sigma^{-1}(f(v_i))$ by running the LM with only the sources specified by $v_i$ and measuring the (sigmoid) probability of $R$.
    \item Solve a Lasso regression problem to find $\hat{w}$ and $\hat{b}$:
    \[
    \hat{w}, \hat{b} \;=\; \arg\min_{w,\,b}\; \frac{1}{n} \sum_{i=1}^n \bigl(g(v_i) - w^\top v_i - b\bigr)^2 \;+\; \lambda \|w\|_1,
    \]
    where $\lambda$ controls sparsity (larger $\lambda$ drives more coefficients to zero).
\end{enumerate}
In \citet{cohen2024contextcite}, typical choices of $n$ range from $32$ to $256$, balancing cost (requires $n$ LM forward passes) and accuracy.
If there are multiple statements $\{r_1, r_2, ..., r_{|R|}\}$ in $R$, the same method can also be applied by focusing only on a subset of tokens in $R$.

\subsection{Heuristic Citation Extraction}
\label{subsec:heuristic}

In our setting, we would like a discrete list of cited sentences for each generated statement, rather than a score for every sentence.
We will now describe how to convert the attribution scores $\hat{w}$ into a discrete subset $C' \subseteq C$ of citations. Let $t$ be a threshold, $p$ be a cumulative probability mass cutoff, and $k$ be a maximum citation limit.

\paragraph{Thresholding and Merging.}
\begin{enumerate}
    \item \textbf{Filtering:} Include only those sources $c_i$ whose attribution score $\hat{w}_i \ge t$. 
    \item \textbf{Merging Adjacent Sources:} If multiple \textit{consecutive} sources in the original text each exceed $t$, merge them into a single “span” $S_j$. We assign this merged span the maximum score among its constituents:
    \[
    \hat{w}(S_j) = \max_{c_i \,\in\, S_j} \hat{w}_i.
    \]
    Here, adjacency is defined by the original ordering in $C$. For instance, if $c_2$ and $c_3$ both pass the threshold and appear consecutively, we merge them into a single span $S_j$.
\end{enumerate}

\paragraph{Softmax Normalization.}
Let $\{S_j\}$ be the set of spans (or single sources) that survived the threshold. We normalize their scores into a probability distribution:
\[
\hat{w}'(S_j) \;=\; \frac{\exp\bigl(\hat{w}(S_j)\bigr)}{\sum_{i} \exp\bigl(\hat{w}(S_i)\bigr)},
\]
so that $\sum_{j} \hat{w}'(S_j) = 1$.

\paragraph{Top-$p$ Selection.}
To avoid including too many low-value sources, we adopt a greedy approach:
\[
\text{Add spans in order of descending } \hat{w}'(S_j)\text{, stopping once } \sum_{S_j \in C'} \hat{w}'(S_j) \,\ge\, p.
\]

\paragraph{Top-$k$ Filtering.}
Finally, if $|C'| > k$, we take only the $k$ highest-scoring spans.

We set $t=1.5$, $p=0.7$, $k=4$ in the experiment.
When generating supervised fine-tuning (SFT) data, we discard any example for which more than 30\% of its statements have no any citations that can survive threshold $t$. This ensures the dataset emphasizes cases where the LM’s response can be tied to explicit context sources. 
We take the LongCite-45K document and question pairs to generate the responses by Llama-3.1-8B-Instruct itself, and then obtain citations with ContextCite (256 calls), transformed into the statement/citation format of LongCite-45K. Finally, we collect $\sim11\text{K}$ examples used for SFT.

\section{Length Balancing}
\label{appx:length}

To prevent the model from simply generating longer citations rather than focusing on citation correctness, we apply a \emph{length balancing} procedure to align the total citation length in our two training responses: a \emph{chosen prediction} and a \emph{reject prediction}. First, we find the citation string (e.g., \texttt{[435-437]}) enclosed in \texttt{<cite>...</cite>} tags for each statement. We then measure each string’s total citation ``coverage'', which means the total number of cited sentences in these intervals.

If a \emph{reject prediction} has a total coverage lower than the corresponding \emph{chosen prediction}, we insert additional citations around nearby sentence indices to match the \emph{chosen} coverage. Conversely, if the \emph{reject} coverage is larger, we randomly remove some of its intervals. We ensure new or inserted citations do not overlap existing intervals and keep them within a small window of 5–10 sentences away from the original citations to maintain realism. Finally, the \emph{reject} and \emph{chosen} will have matched coverage. This approach discourages the model from trivially learning to cite more sentences, instead prompting it to learn \emph{where} and \emph{how} to cite evidence more accurately. Our ablation in Section~\ref{sec:balance} shows that this length balancing technique significantly improves final citation quality.

\section{Comparison with \emph{Claude Citations} API}
\label{appx:claude}

On January 23rd, 2025, Claude announced an API specialized for providing citations along with responses: \emph{Claude Citations}\footnote{\url{https://www.anthropic.com/news/introducing-citations-api}}. We also try to evaluate this API on the LongBench-Cite benchmark. Since the implementation details and resource requirements (e.g., training data) of Claude Citations are not publicly available yet, and it relies on a significantly larger and more powerful LLM, Claude-3.5-Sonnet, which potentially has over 100 billions of parameters, we consider it as a topline of the benchmark rather than a baseline.

When evaluating it on Chinese examples from LongBench-Cite, we found that the API does not split Chinese text properly. As a result, it cites large passages when processing Chinese examples, leading to an average citation length of approximately 800 tokens per citation.

To address this issue, we pre-segment the text ourselves using exactly the same method as our approach following LongCite~\citep{zhang2024longcite}, which uses NLTK and Chinese punctuation segmentation. We then run the Claude Citations API, as it supports both non-segmented and pre-segmented document inputs. The evaluation was conducted using the latest version of \texttt{claude-3-5-sonnet-20241022}.

As shown in Table~\ref{tab:claude_cite}, Claude Citations achieves an overall F1 score of 81.3, which is higher than all other models we have tested.
However, the performance of Claude Citations is not consistent over all datasets. For example, it is worse than \ours on LongBench-Chat and GovReport. The main improvement of Claude is from the DuReader dataset, while the results on other datasets are comparable to the results of \ours. Given the fact that \ours leverages a much smaller 8B model compared to the Claude-3.5-Sonnet model, the result of \ours is very impressive, demonstrating its potential to serve as a strong alternative to proprietary solutions.

\begin{table*}[t]
	\caption{Citation recall (R), citation precision (P), citation F1 (F1), and citation length evaluated on LongBench-Cite benchmark. The best results are bolded. $^\dagger$ indicates the results taken from~\citet{zhang2024longcite}.
	}
	\label{tab:claude_cite}
	\centering
	\small
	\resizebox{\linewidth}{!}{
		\setlength{\tabcolsep}{5pt}
		\begin{tabular}{l|ccc|ccc|ccc|ccc|ccc|c|c}
			\toprule
			\multirow{2}{*}{\bf Model}      &      \multicolumn{3}{c|}{\bf Longbench-Chat}      &      \multicolumn{3}{c|}{\bf MultifieldQA}      &      \multicolumn{3}{c|}{\bf HotpotQA}      &      \multicolumn{3}{c|}{\bf Dureader}      &      \multicolumn{3}{c|}{\bf GovReport}      &      \bf Avg.      &      \bf Citation                                                                               \\ 
			      &      R      &      P      &      F1      &      R      &      P      &      F1      &      R      &      P      &      F1      &      R      &      P      &      F1      &      R      &      P      &      F1      &      \bf F1      &      \bf Length \\ \midrule
			\multicolumn{18}{l}{\textit{\bf Proprietary models}}                                                                                                                                                                                                                                                                                    \\\midrule
			GPT-4o$^\dagger$               &      46.7      &      53.5      &      46.7      &  \bf 79.0      &      87.9      &      80.6      &      55.7      &      62.3      &      53.4      &      65.6      &      74.2      &      67.4      &      73.4      &      90.4      &      79.8      &      65.6      &      220        \\
			Claude-3-sonnet$^\dagger$      &      52.0      &      67.8      &      55.1      &      64.7      &      85.8      &      71.3      &      46.4      &      65.8      &      49.9      &      67.7      &      89.2      &      75.5      &      77.4      &      93.9      &      84.1      &      67.2      &      132        \\
			GLM-4$^\dagger$                &      47.6      &      53.9      &      47.1      &      72.3      &      80.1      &      73.6      &      47.0      &      50.1      &      44.4      &      73.4      &      82.3      &      75.0      &      82.8      &      93.4      &      87.1      &      65.4      &      169        \\ \midrule
			\multicolumn{18}{l}{\textit{\bf Ours: \ours}}      
            \\\midrule

			LongCite-8B (Our repro.)       &      67.0      &      78.1      &      66.6      &      74.8      &      90.7      &      79.9      &      60.8      &      77.9      &      64.1      &      67.1      &      87.2      &      73.7      &      81.6      &      89.3      &      84.5      &      73.8      &      83.5       \\
			+ BoN                          &      68.4      &      81.3      &      71.2      &      76.1      &      92.8      &      81.2      &      67.2      &      81.0      &      68.8      &      70.6      &      90.9      &      76.9      &  \bf 87.6      &      92.4      &  \bf 89.3      &      77.5      &      93.4       \\
			+ SimPO                        &      68.1      &      79.5      &      69.1      &      75.5      &      92.6      &      81.0      &  \bf 69.4      &      82.3      &      71.5      &      72.7      &      91.6      &      78.9      &      86.4      &      92.9      &      89.1      &      77.9      &      105.7      \\
			+ SimPO then BoN               &    \bf 73.3    &      79.4      &    \bf 72.8    &      76.7      &      93.2      &      82.2      &  \bf 69.4      &      83.0      &      71.1      &      74.2      &      92.2      &      80.3      &      86.7      &      92.7      &      89.2      &      79.1      &      94.7       \\

			\midrule
			\multicolumn{18}{l}{\textit{\bf Topline}}                                                                                                                                                                                                                                                            \\
                \midrule
                \it Claude Citations      &      61.2      &      \bf 81.7   &      67.8      &      76.8      &  \bf 98.4      &  \bf 84.9      &      61.9      &  \bf 94.1      &  \bf 72.9      &  \bf 88.5      &  \bf 99.7      &  \bf 93.2      &      79.4      &  \bf 99.2      &      87.7      &  \bf 81.3      &      88.8 \\
			\bottomrule
		\end{tabular}
	}
\end{table*}

\section{Baseline: SimPO with NLI Rewards}
\label{appx:nli}

To provide a stronger fine-tuned baseline, we implement a SimPO variant that adopts NLI-based citation rewards, following the design proposed by ~\citet{huang2024training}. For fair comparison, we keep our full SelfCite SimPO training pipeline—initializing from LongCite-8B and training on the LongCite-45k dataset—and modify only the reward function as a controlled experiment. This NLI-based reward combines two components:

\begin{itemize}
    \item \textbf{Citation Recall Reward:} This measures whether the full set of cited sentences entails the model-generated statement. It is equivalent to the Citation Recall Reward proposed by \citet{huang2024training}.
    
    \item \textbf{Citation Precision Reward:} This estimates whether each cited sentence is necessary by ablating one sentence at a time and testing whether the remaining span still entails the statement. If entailment fails after removing a sentence, it indicates that the sentence contributes uniquely to the justification. To reduce latency, we ablate all sentences when the citation contains 5 or fewer; otherwise, we randomly sample 5 for ablation. When there are N ablations, each ablation makes a reward of $\frac{1}{N}$, and finally all ablations sum up to 1.0. It resembles the Citation Precision Reward proposed by \citet{huang2024training}.
\end{itemize}

We make both rewards positive and capped at 1.0, effectively constructing preference pairs for SimPO. We do not consider the Correctness Recall Reward from \citet{huang2024training}, because the LongCite-45k training set does not contain ground-truth answers. All entailment scores are computed using the public NLI model \texttt{google/t5\_xxl\_true\_nli\_mixture}\footnote{\url{https://huggingface.co/google/t5_xxl_true_nli_mixture}}.

\section{Zero-shot Evaluation on Chunk-level Citation Benchmark ALCE}
\label{appx:alce}

We additionally include the zero-shot evaluation on the chunk-level citation benchmark ALCE~\citep{gao2023enabling} and report the results in Table~\ref{tab:alce}. We find that our baseline model, LongCite-8B, although under a zero-shot setting (it is trained on sentence-level citation but test on chunk-level citations), already outperforms the prompting-based approach from  \citet{gao2023enabling} by a substantial margin in both citation recall and precision. Incorporating NLI-based rewards from \citet{huang2024training} into our SimPO training yields further improvements. Most notably, our method—SimPO with SelfCite rewards—achieves the best performance among models trained on the same LongCite-45k dataset.

The last row of the table presents the best result reported by \citet{huang2024training}, who fine-tuned their model using supervised data. However, this setting is not directly comparable to ours for several reasons:
\begin{enumerate}
    \item They optimize directly for the ALCE evaluation metric by using the same NLI evaluator model (\texttt{google/t5\_xxl\_true\_nli\_mixture}) to provide both training rewards and evaluation scores.
    \item Their model is trained on the \emph{in-distribution} QA training sets in ALCE, with exactly the same chunk-level format as the benchmark. In contrast, our SelfCite model is trained on \emph{out-of-distribution} sentence-level citations from LongCite-45k.
    \item Their method involves distillation from ChatGPT in the first stage, whereas ours does not rely on external supervision.
\end{enumerate}

Despite this domain and format mismatch, SelfCite demonstrates strong generalization and consistently outperforms both LongCite-8B and the NLI-based SimPO baseline. This highlights the robustness and effectiveness of our approach even in cross-domain, cross-format transfer settings.

\begin{table}[t]
\centering
\caption{Evaluation on the chunk-level citation benchmark ALCE~\citep{gao2023enabling}. Our model (SimPO w/ SelfCite) is trained on sentence-level, out-of-distribution LongCite-45k data but still generalizes well to the chunk-level ALCE benchmark.}
\begin{tabular}{l|c|cc|c|cc}
\toprule
\multirow{2}{*}{\textbf{Model}} & \multicolumn{3}{c|}{\textbf{ASQA}} & \multicolumn{3}{c}{\textbf{ELI5}} \\
\cmidrule{2-4}\cmidrule{5-7}
 & \textbf{EM Rec.} & \textbf{Cite Rec.} & \textbf{Cite Prec.} & \textbf{Correct} & \textbf{Cite Rec.} & \textbf{Cite Prec.} \\
\midrule
\multicolumn{7}{c}{\textit{\citet{gao2023enabling} (Prompting)}} \\
\midrule
Llama-2-13B-chat & 34.66 & 37.48 & 39.62 & 12.77 & 17.13 & 17.05 \\
Llama-3.1-8B-Instruct & 42.68 & 50.64 & 53.08 & 13.63 & 34.66 & 32.08 \\
\midrule
\multicolumn{7}{c}{\textit{Finetuned on LongCite-45k (Out-of-Distribution)}} \\
\midrule
LongCite-8B & 42.11 & 62.27 & 57.00 & 15.37 & 30.54 & 29.15 \\
+ SimPO w/ NLI Rewards & 41.20 & 65.65 & 60.20 & 15.30 & 33.06 & 31.05 \\
+ SimPO w/ SelfCite & \textbf{42.57} & \textbf{71.68} & \textbf{62.05} & \textbf{15.17} & \textbf{37.09} & \textbf{35.62} \\
\midrule
\multicolumn{7}{c}{\textit{Finetuned on ALCE train set (In-Distribution Supervision)}} \\
\midrule
\citet{huang2024training} & 40.05 & 77.83 & 76.33 & 11.54 & 60.86 & 60.23 \\
\bottomrule
\end{tabular}
\label{tab:alce}
\end{table}

\section{Comparison with Prior Studies}
\label{appx:comparison}

We further provide a comparison table in Table~\ref{tab:method-comparison} to contrast the key differences between SelfCite and other prior studies on producing citations from LLMs. Among all methods, SelfCite is the only approach that supports sentence-level citation generation in a single pass, leverages preference optimization, and scales to 128K-token contexts—all without requiring additional supervision. In contrast, prior work such as ALCE~\citep{gao2023enabling} and ~\citet{huang2024training} use chunk-level citations for shorter context ($\leq$8K) and require prompt-based or supervised NLI signals. ContextCite~\citep{cohen2024contextcite}, while being sentence-level, relies on a computationally expensive (at least 32 inference calls) process for random context ablation and trains a linear model for estimating the importance scores.
This comparison underscores the practical advantages and technical contributions of SelfCite.

\begin{table}[ht]
\centering
\small
\caption{Key differences among prior methods on producing citations from LLMs. CC stands for ContextCite.}
\begin{tabular}{lccccc}
\toprule
\bf Method &
\shortstack{Sentence‐level\\ citations?} &
\shortstack{One pass\\ generation?} &
\shortstack{Preference\\ optimization?} &
\shortstack{Handle 128K\\ long‐context?} &
\shortstack{External\\ supervision?} \\
\midrule
ALCE~\citep{gao2023enabling}                    & \xmark{} (chunk-level)        & \cmark{}                     & \xmark{} (prompting)            & \xmark{} (8K)             & 2‑shot prompting         \\
\citet{huang2024training}      & \xmark{} (chunk-level)        & \cmark{}                     & \cmark{}                        & \xmark{} (8K)             & NLI + ground truth       \\
CC~\citep{cohen2024contextcite}             & \cmark{}                      & \xmark{} (at least 32 calls) & \xmark{} (not generative)       & \cmark{}                  & N/A                      \\
LongCite~\citep{zhang2024longcite}                & \cmark{}                      & \cmark{}                     & \xmark{} (SFT only)             & \cmark{}                  & SFT data                 \\
SelfCite (Ours)         & \cmark{}                      & \cmark{}                     & \cmark{}                        & \cmark{}                  & N/A                      \\
\bottomrule
\end{tabular}
\label{tab:method-comparison}
\end{table}

\section{More Qualitative Examples}
\label{appx:qual}

We further show more qualitative examples in Table~\ref{tab:q1},\ref{tab:q2}, and \ref{tab:q3}, to represent the cases where \ours is better as well as where the LongCite-8B direct sampling baseline is better. In Table~\ref{tab:q1}, \ours BoN avoid the cited irrelevant sentence (42, 47-50) by the baseline, while further including a correct citation (23) that are not found by the baseline. In Table~\ref{tab:q2}, both \ours BoN and the baseline cites too many irrelevant sentences (391-393) but \ours BoN's citation is slightly better. In Table~\ref{tab:q3}, \ours BoN wrongly includes 30 and misses 70, but the baseline is slightly better and only wrongly includes 71.

\begin{table*}[t!]
    \vskip -0.1 in
\caption{An example of differences in the citation from baseline vs BoN. Related information are highlighted in the context/response.}
\label{tab:q1}
    \vskip 0.05 in
\centering
\small
\renewcommand{\arraystretch}{1.0} 
\begin{tabular}{p{0.1\textwidth} p{0.85\textwidth}}
\toprule
\bf Sent. ID & \textbf{Context Sentences} (only showing cited sentences due to space) \\
\midrule
\textbf{23 (\ding{51})} 
  & We explored using \hl{a simple and cost-effective procedure to instruction finetune our continually pretrained long models without any human-annotated data.}\\
\midrule
\textbf{42 (\ding{55})} 
  & Collecting human demonstration and preference labels for LLM alignment is a cumbersome and expensive process (Ouyang et al., 2022; Touvron et al., 2023).\\
\midrule
\textbf{45 (\ding{51})} 
  & In this work, we found that a simple and cheap approach which leverages \hl{a pre-built large and diverse short-prompt dataset} works surprisingly well on long-context benchmarks.\\
\midrule
\textbf{46 (\ding{51})} 
  & Specifically, we take the RLHF dataset used in LLAMA 2 CHAT and \hl{augment it with synthetic self-instruct} (Wang et al., 2022) \hl{long data generated by LLAMA 2 CHAT itself}, in the hope that the model can learn a diverse set of skills through the large amount of RLHF data and transfer that knowledge to long-context scenarios via self-instruct data.\\
\midrule
\textbf{47-50 (\ding{55})} 
  & The data generation process focuses on QA-format tasks: starting from a long document in our pretraining corpus, we select a random chunk and prompt LLAMA 2 CHAT to write question-answer pairs based on information in the text chunk. [...]\\
\midrule
\midrule
\bf Query & What aspects of the LLAMA Long model proposed above have changed relative to the LLAMA-based model? What improvements have been made? \\
\midrule
\textbf{Response} (single statement due to space)
 & [...] 3. Instruction Tuning: The paper proposes \hl{a simple and cost-effective procedure to instruction finetune the continually pretrained long models without any human-annotated data.} This involves using \hl{a pre-built large and diverse short-prompt dataset} and \hl{augmenting it with synthetic self-instruct long data generated by LLAMA CHAT itself.} [...]\\
\midrule
\multicolumn{2}{l}{\textbf{Citation Strings (\textcolor{ForestGreen}{\it green: correct}; \textcolor{red}{\it red: wrong})}} \\
\midrule
\textbf{Baseline} 
 & \texttt{\textcolor{red}{[42-42]}\textcolor{ForestGreen}{[45}\textcolor{red}{-50]}} \\
\textbf{\ours BoN} 
 & \texttt{\textcolor{ForestGreen}{[23-23][45-45][46-46]}} \\
\midrule
\end{tabular}
\end{table*}
\begin{table*}[t!]
    \vskip -0.1 in
\caption{An example of differences in the citation from baseline vs BoN. Related information are highlighted in the context/response.}
\label{tab:q2}
    \vskip 0.05 in
\centering
\small
\renewcommand{\arraystretch}{1.0} 
\begin{tabular}{p{0.1\textwidth} p{0.85\textwidth}}
\toprule
\bf Sent. ID & \textbf{Context Sentences} (only showing cited sentences due to space) \\
\midrule
\textbf{299 (\ding{55})} 
  & Ribosomes link amino acids together in the order specified by the codons of messenger RNA (mRNA) molecules to form polypeptide chains.\\
\midrule
\textbf{300 (\ding{51})} 
  & \hl{Ribosomes consist of two major components: the small and large ribosomal subunits.}\\
\midrule
\textbf{301 (\ding{51})} 
  & \hl{Each subunit consists of one or more ribosomal RNA (rRNA) molecules and many ribosomal proteins (RPs or r-proteins).}\\
\midrule
\textbf{302 (\ding{51})} 
  & \hl{The ribosomes and associated molecules are also known as the translational apparatus.}\\
\midrule
\textbf{383 (\ding{55})} 
  & Ribosomes are minute particles consisting of RNA and associated proteins that function to synthesize proteins. \\
  \midrule
\textbf{384 (\ding{55})} 
  & Proteins are needed for many cellular functions such as repairing damage or directing chemical processes. \\
  \midrule
\textbf{385 (\ding{55})} 
  & Ribosomes can be found floating within the cytoplasm or attached to the endoplasmic reticulum.\\
\midrule
\textbf{390 (\ding{51})} 
  & \hl{Each subunit is made of one or more rRNAs and many r-proteins.}\\
\midrule
\textbf{391 (\ding{55})} 
  & The small subunit (30S in bacteria and archaea, 40S in eukaryotes) has the decoding function, whereas the large subunit (50S in bacteria and archaea, 60S in eukaryotes) catalyzes the formation of peptide bonds, referred to as the peptidyl-transferase activity.\\
\midrule
\textbf{392 (\ding{55})} 
  & The bacterial (and archaeal) small subunit contains the 16S rRNA and 21 r-proteins (Escherichia coli), whereas the eukaryotic small subunit contains the 18S rRNA and 32 r-proteins (Saccharomyces cerevisiae; although the numbers vary between species).\\
\midrule
\textbf{393 (\ding{55})} 
  & The bacterial large subunit contains the 5S and 23S rRNAs and 34 r-proteins (E. coli), with the eukaryotic large subunit containing the 5S, 5.8S, and 25S / 28S rRNAs and 46 r-proteins (S. cerevisiae; again, the exact numbers vary between species).\\
\midrule
\midrule
\bf Query & The large subunit and small subunit that use two types of RNA are major components that make up what? \\
\midrule
\textbf{Response} (single statement due to space)
 & [...] The ribosome consists of \hl{two major components: the small and large ribosomal subunits.} \hl{Each subunit consists of one or more ribosomal RNA (rRNA) molecules and many ribosomal proteins (RPs or r-proteins).} \hl{The ribosomes and associated molecules are also known as the translational apparatus}. [...]\\
\midrule
\multicolumn{2}{l}{\textbf{Citation Strings (\textcolor{ForestGreen}{\it green: correct}; \textcolor{red}{\it red: wrong})}} \\
\midrule
\textbf{Baseline} 
 & \texttt{\textcolor{red}{[299-}\textcolor{ForestGreen}{302]}\textcolor{red}{[383-385]}\textcolor{ForestGreen}{[390}\textcolor{red}{-393]}} \\
\textbf{\ours BoN} 
 & \texttt{\textcolor{ForestGreen}{[300-302]}\textcolor{ForestGreen}{[390}\textcolor{red}{-393]}} \\
\midrule
\end{tabular}
\end{table*}
\begin{table*}[t!]
    \vskip -0.1 in
\caption{An example of differences in the citation from baseline vs BoN. Related information are highlighted in the context/response.}
\label{tab:q3}
    \vskip 0.05 in
\centering
\small
\renewcommand{\arraystretch}{1.0} 
\begin{tabular}{p{0.1\textwidth} p{0.85\textwidth}}
\toprule
\bf Sent. ID & \textbf{Context Sentences} (only showing cited sentences due to space) \\
\midrule
\textbf{28 (\ding{51})} 
  & \hl{The Aegis BMD system exists in several variants.} \\
\midrule
\textbf{29 (\ding{51})} 
  & Listed in order of \hl{increasing capability}, these include (but are not necessarily limited to) \hl{3.6.X variant, the 4.0.3 variant, the 4.1 variant} (also known as the Aegis Baseline [BL] 5.4 variant), the \hl{5.0 CU} (Capability Upgrade) variant (also known as the BL 9.1 variant), the \hl{5.1 variant} (also known as the BL 9.2 variant), and the \hl{6.X variant} (also known as the BL 10 variant). \\
\midrule
\textbf{30 (\ding{55})} 
  & Figure 1 summarizes the capabilities of some of these variants (using their designations as of 2016) and correlates them with the phases of the European Phased Adaptive Approach (or EPAA; see discussion below) for European BMD operations.\\
\midrule
\textbf{61 (\ding{51})} 
  & The BMD \hl{interceptor missiles used by Aegis ships are the Standard Missile-3 (SM-3), the SM-2 Block IV, and the SM-6.} \\
\midrule
\textbf{62 (\ding{51})} 
  & \hl{The SM-3 is designed to intercept ballistic missiles above the atmosphere} (i.e., exo-atmospheric intercept), in the midcourse phase of an enemy ballistic missile's flight.\\
\midrule
\textbf{70 (\ding{51})} 
  & \hl{The SM-2 Block IV is designed to intercept ballistic missiles inside the atmosphere} (i.e., endo-atmospheric intercept), during the terminal phase of an enemy ballistic missile's flight.\\
\midrule
\textbf{71 (\ding{55})} 
  & It is equipped with a blast fragmentation warhead.\\
\midrule
\midrule
\bf Query & Please write a one-page summary of the above government report. \\
\midrule
\textbf{Response} (single statement due to space)
 & [...] \hl{The Aegis BMD system exists in several variants} with increasing capability, including \hl{the 3.6.X, 4.0.3, 4.1, 5.0 CU, 5.1, and 6.X variants}. The system uses the \hl{Standard Missile-3 (SM-3), SM-2 Block IV, and SM-6 interceptor missiles}. The SM-3 is designed to \hl{intercept ballistic missiles above the atmosphere in the midcourse phase}, while the SM-2 Block IV and SM-6 are designed to intercept ballistic missiles inside the atmosphere in the terminal phase. [...]\\
\midrule
\multicolumn{2}{l}{\textbf{Citation Strings (\textcolor{ForestGreen}{\it green: correct}; \textcolor{red}{\it red: wrong})}} \\
\midrule
\textbf{Baseline} 
 & \texttt{\textcolor{ForestGreen}{[28-29][61-62][70-}\textcolor{red}{71]}} \\
\textbf{\ours BoN} 
 & \texttt{\textcolor{ForestGreen}{[28-}\textcolor{red}{30]}\textcolor{ForestGreen}{[61-62]}} \textcolor{red}{(missing: 70)}\\
\midrule
\end{tabular}
\end{table*}

\end{document}